# A Comparative Study of 3D Model Acquisition Methods for Synthetic Data Generation of Agricultural Products

Steven Moonen[1][0000-0002-0935-701X] and Rob Salaets[2][0000-0002-4835-0793] and Kenneth Batstone[2][0000-0001-8328-1052] and Abdellatif Bey-Temsamani[2][0000-0002-8198-0965] and Nick Michiels[1][0000-0002-7047-5867]

[1] Hasselt University - Flanders Make, Digital Future Lab, Diepenbeek 3590, Belgium
[2] Flanders Make, Gaston Geenslaan 8, Heverlee 3001, Belgium
`steven.moonen@uhasselt.be`

**Abstract.** In the manufacturing industry, computer vision systems based on artificial intelligence (AI) are widely used to reduce costs and increase production. Training these AI models requires a large amount of training data that is costly to acquire and annotate, especially in high-variance, low-volume manufacturing environments. A popular approach to reduce the need for real data is the use of synthetic data that is generated by leveraging computer-aided design (CAD) models available in the industry. However, in the agricultural industry these models are not readily available, increasing the difficulty in leveraging synthetic data. In this paper, we present different techniques for substituting CAD files to create synthetic datasets. We measure their relative performance when used to train an AI object detection model to separate stones and potatoes in a bin picking environment. We demonstrate that using highly representative 3D models acquired by scanning or using image-to-3D approaches can be used to generate synthetic data for training object detection models. Finetuning on a small real dataset can significantly improve the performance of the models and even get similar performance when less representative models are used.

**Keywords:** Synthetic data, Computer vision, Agriculture.

## 1 Introduction

Computer vision systems are widely used in manufacturing for sorting and inspecting products [1-4], reducing costs by automating production and detecting faults early. These systems rely on AI models trained on large, annotated datasets, which are expensive to create. Research explores replacing parts of these datasets with synthetic data [5, 8]. Typically, CAD models are used to render photorealistic images while exporting labels automatically. In the manufacturing industry, CAD models are created in the design phase, making them readily available for synthetic data generation. In agriculture, however, no digital 3D models of the products exist, and creating them from scratch is costly and have to be recreated when new products are cultivated. A 3D artist must model and texture multiple objects with high variety to prevent overfitting since the shape of the products are not static as is the case in the manufacturing industry. Alternatively, the artist can leverage procedural techniques using noise and randomization to generate variations.

These techniques required skilled artists and time. To reduce costs other techniques to create 3D models are explored. Generative AI can create 3D models from text or images, but these methods have limitations. Text prompts offer little control, and image-based approaches struggle to reconstruct unseen parts of objects while also embedding lighting present in the scene into the textures. Photogrammetry reconstructs 3D models from images taken from multiple viewpoints, while 3D scanners provide high accuracy but require expensive equipment. With proper calibration, these techniques can remove environmental lighting from the textures, making them suitable for new digital lighting conditions.

Albayrak et al. [9] did a similar experiment for the manufacturing industry, focusing on different scanning methods and varying levels of detail in the modeling. Gao et al. [10] used 2D masks to cut out and combine weeds from multiple images into new synthetic images. This technique circumvents the need for 3D models and reduces the domain gap but requires pixel perfect annotations.

## 2 Method

This paper will focus on a sorting use case, where potatoes and stones are present in Euronorm containers. The goal is to have a computer vision system that can classify objects inside the box between potatoes and stones. To achieve this, we train an AI model on synthetic data generated with CAD2Render [11].

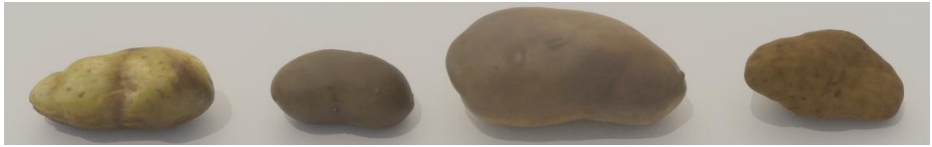

**Fig. 1.** Renders of the potato models in CAD2Render. The 3D models were acquired with different techniques from left to right: text prompt to 3D using Meshy, image to 3D using Trellis, Scanned using an Einscanner HX, procedurally generated using Blender.

### 2.1 3D model acquisition

We created the 3D models with four different approaches. With each approach, we created six 3D models for the potatoes and five for the stones. Each modeling technique also created albedo textures, all other material properties are set to be identical across all models of the same class. In Figure 1, examples for each model technique are visible. For both the models that are scanned and generated from images, the exact same real potatoes and stones were used. The potatoes were acquired from a single batch at a local supermarket, while the stones were acquired from a local public park.

- **Text-to-3D:** We used Meshy [12], a generative AI application that uses a text prompt to generate 3D models. The text prompt used for generating the assets where: "A freshly harvested potato, photo realistic, 3D scan" and "A freshly dug up rock, photo realistic, 3D scan". From the results, the most representative models were selected. Both the mesh generation and texture generation take about half a minute to compute. Combined with the prompt engineering and selecting the most representative models, this takes about 2 minutes for each 3D model that is generated.

- **Image-to-3D:** We used Trellis [13], an image-to-3D AI application that supports both single and multi-view input images for generating a single 3D model. Using only a single image often results in inaccurate or heavy discolorations on the bottom or far side of the generated models. Therefore, we used 6 images for each generation. 3 images were taken evenly spaced around the object before turning it over and repeating this process. The images where taken with a clear white background and diffuse light to minimize baking the environment light inside the generated albedo textures. These textures where then further delit with Agisoft Texture DeLighter [14]. After the initial setup, it takes 3 minutes to generate the 3D model.
- **Scanning:** We used the Einscanner HX to scan accurate 3D models. The objects were scanned in two passes, turning the object over in-between passes. The two scans are then combined using the software provided with the scanner. This process takes about 30 minutes to complete and requires dedicated hardware.
- **3D modeling:** We used the Python scripting API and shading nodes of Blender [15] to generate potato models and textures. Starting from an icosphere, multiple modifiers are applied to create a plausible potato shape. The texture is then generated through a series of noise patterns and texture modifiers. All scripts used for generating the models and textures are available on the GitHub page [11]. Once the necessary scripts are created, new 3D models can be generated in seconds. Preparing the scripts can take a full work day or even longer for complex 3D models.

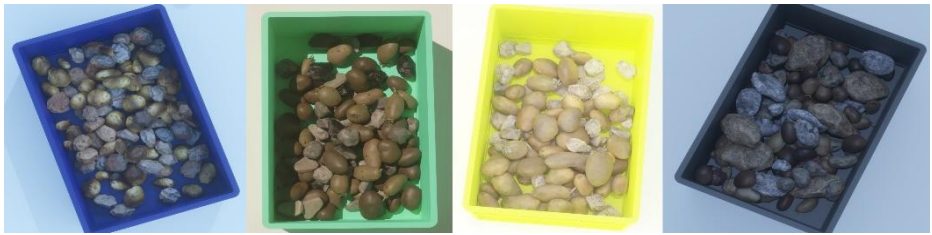

**Fig. 2.** Renders of the synthetic images generated with CAD2Render. The 3D models used in these images were acquired with different techniques from left to right: Text-to-3D using Meshy, Image-to-3D using Trellis, Scanned using an Einscanner HX, modelled using Blender.

### 2.2 Data generation

To create a synthetic dataset the CAD2Render toolkit [11] was used. This Unity toolkit randomizes parameters to boost domain randomization and reduce performance loss in training vision AI on synthetic data. The parameters that we randomize are: environment light, white balancing, exposure, small color variation of the potato textures, scale of the models, colors of surrounding scene objects.

For each technique used to acquire the 3D models described in Section 2.1, a new dataset is generated while keeping all other parameters constant. Each dataset consists of 1000 images where every stone and potato is annotated conform to the BOP format [16]. The complete scene used to generate the data is available on the CAD2Render GitHub page [11]. A sample image for each dataset is visible in Figure 2.

## 2.3 Real data acquisition

The real data consists of images taken with a Lucid Vision Labs 3.2MP TRI032S-C camera on a conveyor system at a resolution of 1873 x 1340 pixels. The scene consists of a mix of potatoes with stones, small lumps of dirt, small leaves and twigs in four different colored Euronorm food containers. All items in the containers are randomly placed by a human. The potatoes and stones were acquired from a local farm. For each container an example image of the dataset is displayed in Figure 3. In total the dataset consists of 50 images with 1784 potatoes and 229 stones. The dataset is manually labeled with instance segmentation masks and bounding boxes.

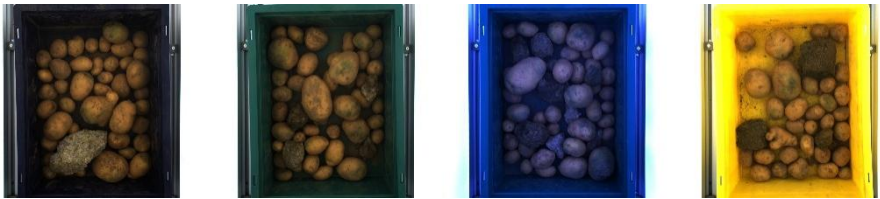

**Fig. 3.** Examples of the real images of the potatoes and stones used in the test set.

## 2.4 AI model training

To keep the focus on dataset quality, we employ a standard object detection pipeline. A YOLOv8n model architecture [17], this architecture has 3 million parameters and requires 8.2 GFLOPs for evaluation. The longest edge of the images is resized to 640 pixels. Five default data augmentation techniques were used during the training: HSV-color space distortion, panning/padding, scaling, flipping, mosaic augmentation.

A new model was pretrained on each of the synthetic datasets described in section 2.2. The datasets were split with 900 images for training and 100 images for validation. Early stopping is used with a patience of 100 epochs to make sure convergence is reached. Fine-tuning on real data is done afterwards with a warmup period of three epochs. For the fine-tuning 4 random images were taken from the real image dataset making sure one image of each colored Euronorm container was selected. The pretraining takes ~3200s and fine-tuning ~200s on a NVIDIA GeForce RTX 2080 Ti.

## 3 Results

We tested our models on the real data excluding any images used during the fine-tuning of the models for a total of 46 images. We use the mean average precision (mAP50) metric for evaluation. The mAP50 metric is a commonly used metric for object detection that measures the average precision across multiple classes at a fixed Intersection over Union threshold of 50%. It quantifies the model's ability to correctly detect and classify objects by computing the area under the precision-recall curve for each class and averaging the values across all classes. The results of these tests are listed in Table 1.

**Table 1.** Table containing the Mean average precision and recall at confidence 50% for each model (YOLOv8n) trained on the synthetic datasets. The synthetic datasets are created with different techniques of acquiring the 3D models. All tests were performed on 46 real images. The finetuned models use 4 real images after pretraining on the synthetic datasets. The last row displays the results of a model with randomly initialized weights instead of being pretrained.

| Dataset | mAP50 | mAP50 Potato | mAP50 Stone | Recall Potato | Recall Stone |
| --- | --- | --- | --- | --- | --- |
| Text-to-3D | 0.314 | 0.568 | 0.059 | 0.439 | 0.444 |
| Image-to-3D | **0.676** | **0.926** | **0.425** | **0.756** | 0.569 |
| Scanned | 0.661 | 0.924 | 0.398 | **0.756** | 0.583 |
| 3D Modeled | 0.519 | 0.804 | 0.233 | 0.622 | **0.759** |
| Text-to-3D finetuned | 0.945 | 0.981 | 0.909 | 0.957 | 0.799 |
| Image-to-3D finetuned | **0.956** | **0.985** | **0.927** | **0.961** | **0.870** |
| Scanned finetuned | 0.942 | 0.983 | 0.900 | 0.946 | 0.787 |
| 3D Modeled finetuned | 0.946 | 0.984 | 0.909 | 0.951 | 0.759 |
| Random init. finetuned | 0.781 | 0.942 | 0.621 | 0.920 | 0.587 |

## 4 Conclusion

After pretraining with any of the synthetic datasets, the AI models can detect most objects (stones and potatoes) in the images. However, classification accuracy declines when the 3D models are less representative of the real objects. Especially the Text-to-3D method scores significantly lower than all other methods as this produces the least representative models. All other techniques use real example objects as some form of input to guide the modeling. The manually modeled approach slightly under performs to the scanned and Image-to-3D techniques.

Training solely on synthetic data can yield good detection and classification results when models are highly representative. However, all techniques benefit from using even a small real-image dataset for finetuning. With finetuning the differences in performance between the different techniques become insignificant, only the Image-to-3D has a noticeable improvement in the recall of the stone detection. However, the results are insufficient to establish Image-to-3D as a more effective 3D acquisition method for performance reasons. Nevertheless, it is also one of the fastest methods discussed in this paper making it the most relevant in most use cases.

Even though the different acquisition techniques do not have any meaningful differences between each other after finetuning, they still give a clear benefit compared to no pretraining at all. Highlighting the benefits of pretraining on synthetic data.

**Acknowledgments:** This research was realized in the framework of the NORM.AI SBO project, funded by Flanders Make, the strategic research Centre for the Manufacturing Industry in Belgium. This work was made possible with support from MAXVR-INFRA, a scalable and flexible infrastructure that facilitates the transition to digital-physical work environments.